\def\year{2018}\relax
\documentclass[letterpaper]{article} %DO NOT CHANGE THIS
\usepackage{aaai18}  %Required
\usepackage{times}  %Required
\usepackage{helvet}  %Required
\usepackage{courier}  %Required
\usepackage{url}  %Required
\usepackage{graphicx}  %Required
\usepackage{booktabs} % For formal tables
\usepackage{bm}
\usepackage{graphicx}
\usepackage{subfig}
\usepackage{amsmath}
\usepackage[noend]{algpseudocode}
\usepackage{algorithmicx,algorithm}
\usepackage{makecell}  
\usepackage{multirow,array,tabularx}
\usepackage{amsfonts}
\makeatletter
\def\hlinew#1{%
  \noalign{\ifnum0=`}\fi\hrule \@height #1 \futurelet
   \reserved@a\@xhline}
\makeatother

\begin{document}
% The file aaai.sty is the style file for AAAI Press 
% proceedings, working notes, and technical reports.
%
\title{Harmonic Adversarial Attack Method}
\author{
  Wen Heng\textsuperscript{1,2},
  Shuchang Zhou\textsuperscript{1},
  Tingting Jiang\textsuperscript{2},
  \\ %\footnotemark[1]
\textsuperscript{1}Megvii Inc(Face++)
\textsuperscript{2}Peking University\\
{wenheng@pku.edu.cn, zsc@megvii.com, ttjiang@pku.edu.cn}}
\maketitle
\begin{abstract}
Adversarial attacks find perturbations that can fool models into misclassifying images. Previous works had successes in generating noisy/edge-rich adversarial perturbations, at the cost of degradation of image quality. Such perturbations, even when they are small in scale, are usually easily spottable by human vision. In contrast, we propose Harmonic Adversarial Attack Methods (HAAM), that generates edge-free perturbations by using harmonic functions. The property of edge-free guarantees that the generated adversarial images can still preserve visual quality, even when perturbations are of large magnitudes. Experiments also show that adversaries generated by HAAM often have higher rates of success when transferring between models. In addition, we find harmonic perturbations can simulate natural phenomena like natural lighting and shadows. It would then be possible to help find corner cases for given models, as a first step to improving them.
\end{abstract}

\section{Introduction}
% \begin{figure}[t]
% \centering
% \subfloat[(a) True label: megalithic]{
% \includegraphics[width=0.24\textwidth]{fig/ref.png}}\\
% \subfloat[(b) Predicted label: chain]{
% \includegraphics[width=0.24\textwidth]{fig/haam_.png}}
% \subfloat[(c) Harmonic perturbation]{
% \includegraphics[width=0.24\textwidth]{fig/haam_e.png}}\\
% \subfloat[(d) Predicted label: chain]{
% \includegraphics[width=0.24\textwidth]{fig/fgsm_.png}}
% \subfloat[(e) Noise perturbation]{
% \includegraphics[width=0.24\textwidth]{fig/fgsm_e.png}}
% \caption{(a) Reference image. (b) Adversarial image generated using HAAM with $\epsilon = 6$. (c) Harmonic perturbation. (d) Adversarial image generated using FGSM with $\epsilon=8$. (e) Noise distortion. The adversarial image generated using HAAM is indistinguishable from natural images. }
% \label{fig:adv_examples}
% \end{figure}

\begin{figure}[t]
\centering
\includegraphics[width=0.5\textwidth]{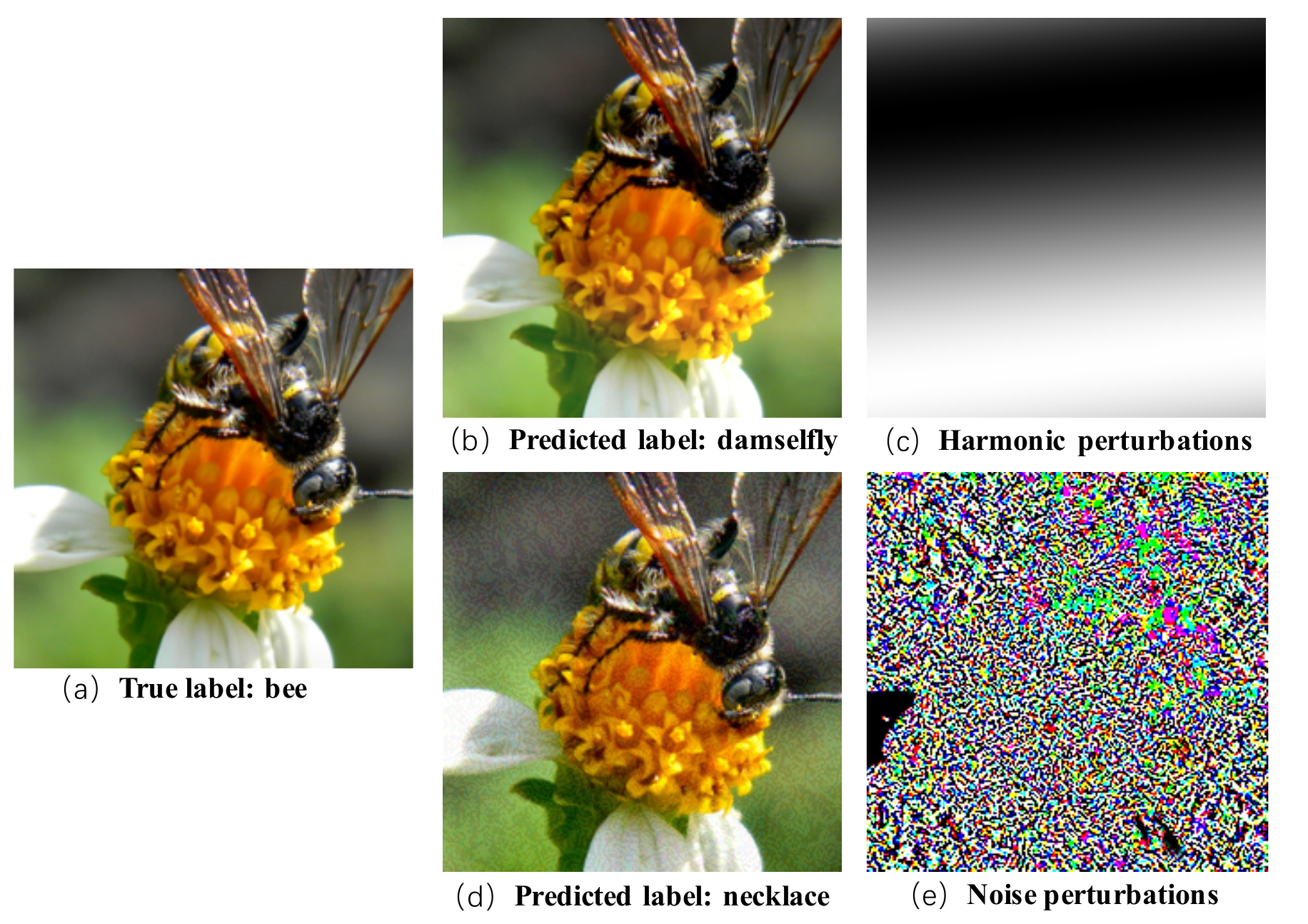}
\caption{(a) Reference image. (b) Adversarial image generated using HAAM with $\epsilon = 24$. (c) Harmonic perturbations of (b). (d) Adversarial image generated using FGSM with $\epsilon=8$. (e) Noisy perturbations of (d). The adversarial image generated using HAAM is indistinguishable from natural images. }
\label{fig:adv_examples}
\end{figure}

Deep neural networks (DNNs) have made great progresses in a variety of application domains, like computer vision, speech and many other tasks~\cite{krizhevsky2012,vgg,resnet,hinton2012deep,clark2015training,socher2011parsing}. However, it has been shown by many works that the state-of-the-art DNNs are vulnerable to adversarial examples, which are images generated by adding carefully designed perturbations on the natural images~\cite{FGSM,BIM,transferability,deepfool,fawzi2018analysis}. The adversarial attacks reveal the weakness of DNNs models, even though they achieve human-competitive performances in many tasks. More importantly, adversarial examples also pose potential security threats to machine learning systems in practice, and may stunt the growth of applying DNNs in practice. Therefore, the study of adversarial attacks is crucial to improving the robustness of models.

\begin{figure*}[t]
\centering
\includegraphics[width=0.98\textwidth]{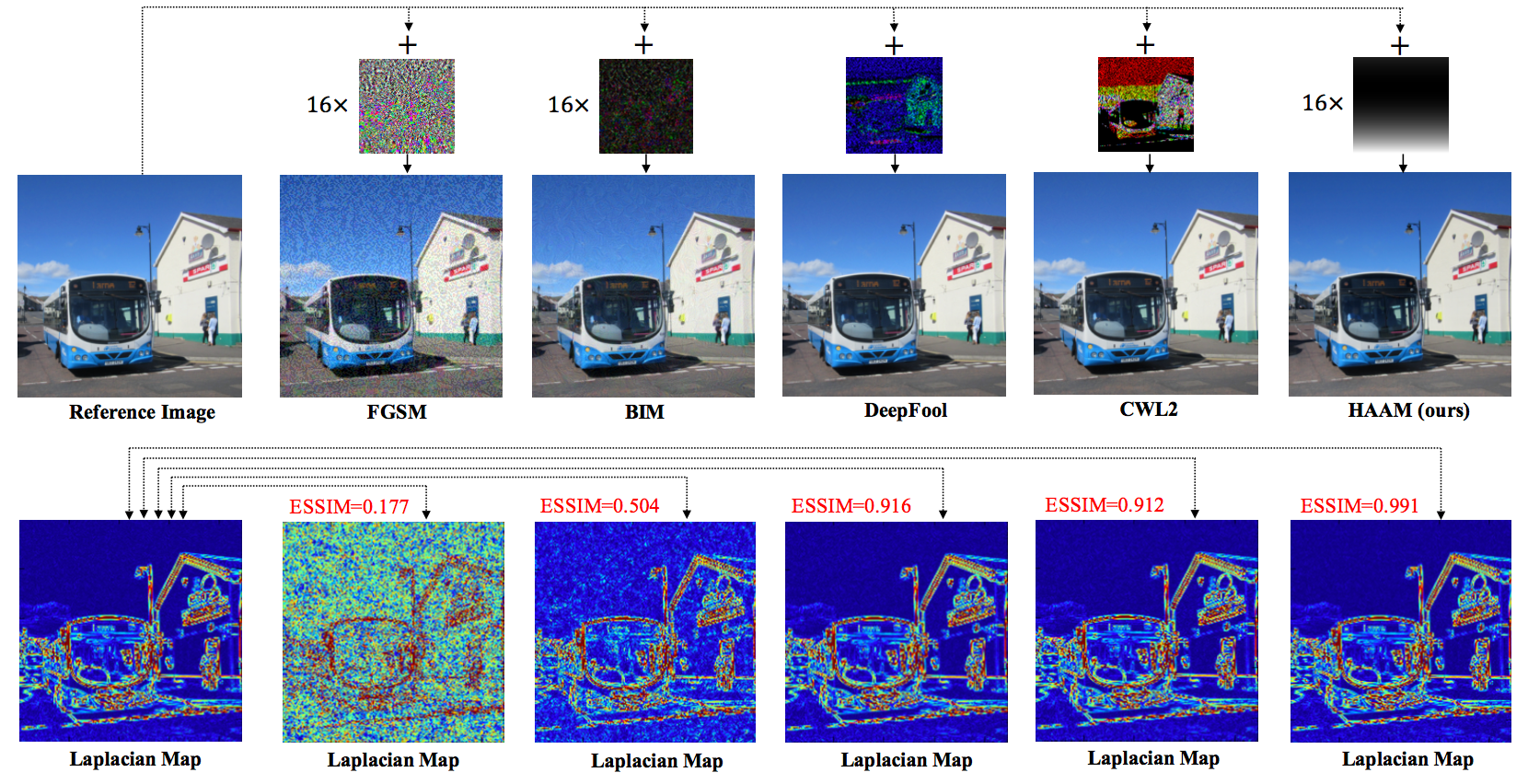}
\caption{First row shows the reference and its adversarial images generated by five adversarial methods. Second row shows their corresponding laplacian maps. We calculated the Edge-SSIM (ESSIM) between the reference and distorted laplacian maps. It's shown that perturbations generated by HAAM don't disturb the laplacian maps and adversaries show good visual quality. ESSIM is to apply SSIM on the laplacian maps. (Zoom in to see details)}
\label{fig:lap_cmp}
\end{figure*}

Most of existing adversarial methods~\cite{FGSM,BIM,deepfool,cwl2} generate pixel-wise perturbations of limited magnitude. The perturbations usually show random patterns that are rich in edges, as is shown in Fig~\ref{fig:adv_examples} (e). Such perturbations inevitably change the spatial frequency of natural images. Human vision is quite sensitive to the edge information, as there is a primary visual cortex (V1) which is devoted to edge extraction~\cite{v1_cortex}. Hence the adversarial patterns generated by adding pixel-wise noisy perturbations tend to be easily spottable by human vision. In addition, to reduce the magnitude of perturbations and meanwhile achieving high attacking rate, carefully designed adversarial methods usually only work effectively when models are known (white-box attack), but the adversarials may not transfer to unknown models (black-box attack)~\cite{adv_scale}. On the other hand, adversarial examples generated by Fast Gradient Sign Method (FGSM)~\cite{FGSM} have a good transferability, since FGSM is an one-step attack method which generate noisy/edge-rich adversarial images, but suffer in visual quality. In short, high visual quality and transferability are hard to be achieved at the same time. 

We propose Harmonic Adversarial Attack Method (HAAM), a novel adversarial method to narrow the gap between visual quality and transferability. Different from previous noisy/edge-rich perturbations, the proposed method can generate edge-free perturbations, by requiring the laplacian of generated perturbations to be equal to 0 at any point. Consequently, there would be no detectable edges in the perturbations~\cite{shapley1973edge}. We enforce the constraint by using harmonic functions as parametric models to generate perturbations, since harmonic functions satisfy the Laplace's equations~\cite{math_physics} and can be constructed from analytic complex functions. In experiments, we find that the adversarial images produced by HAMM are still of good visual quality, even when the magnitudes of added perturbations are quite large. In addition, the large magnitudes of perturbations guarantee good transferability between models. Moreover, by using special harmonic functions, the generated adversarial images may look like some natural phenomena. This can be used to generate corner cases for target models and reveal their weakness in nature environments.

In summary, our contributions are as follows.
\begin{itemize}
\item We propose an adversarial method which can generate edge-free adversarial perturbations. In experiments, we find the adversarial examples can strike a balance between visual quality and transferability.
\item We propose using analytic complex functions as parametric models to systematically construct harmonic functions. The parameters are learned in an end-to-end fashion.
\item With special harmonic functions, we find the generated adversarial examples can simulate some natural phenomena and/or photographic effects. This helps to find real-life corner cases for target models and give suggestions for improving the models.
\end{itemize}
\section{Related Works}
The research of adversarial attacks on neural networks has made great progresses in recent years. Fast Gradient Sign Method (FGSM)~\cite{FGSM} is a one-step attack method, which shifts the input by $\epsilon$ in the direction of minimizing the adversarial loss. Based on FGSM, Basic Iterative Method (BIM) was proposed ~\cite{BIM}. Compared to FGSM, adversaries generated by BIM show less perturbations and consequently have a higher success rate in attacking. DeepFool~\cite{deepfool} is a powerful attack method, which shifts the input with least perturbations in the direction to its nearest classification plane. It's shown that the adversaries generated by DeepFool have sparser perturbations, compared to FGSM and BIM. In addition, C\&W ~\cite{cwl2} was proposed based on three distance metrics: $L_0$, $L_2$ and $L_{\infty}$. With carefully designed optimization loss for perturbations searching, the C\&W methods are shown to successfully overtake the defensive distillation~\cite{distillation} with 100\% success rate. In summary, most of existing adversarial methods pursue a high success rate in attacking, but ignore the natural properties of adversarial images. Since the perturbations generated by these methods are usually noisy/edge-rich, this will make the adversarial images look unnatural in some degree. More importantly, such images barely exist in the real world. 

Different from the methods which generate noisy/edge-rich perturbations, some works focus on generating physical-world adversaries. This kind of adversaries can attack target models successfully in the physical environment. ~\cite{sunglass_adv} proposes generating adversarial perturbations on the sunglasses mask which can be printed out and weared by people to trick the face recognizers. In ~\cite{adv_patch}, the authors generate a universal sticker, which can make any object recognized as a `toaster' by the neural network classifier. Similarly, in~\cite{print3d_adv} the authors take 3D-printing technique to build one adversarial object. The object would be always recognized as the target class no matter with which angle it was captured by cameras. ~\cite{trafficsign_adv} proposes generating subtle posters or stickers which can be posted on traffic signs to cheat the traffic signs recognizer, which are crucial to a Autonomous Driving system. %This shows that current DNNs used in practical applications, are rather vulnerable to attacks. 
~\cite{natural_adv} try to generate more natural adversaries using Generative Adversarial Networks (GAN)~\cite{gan}.
However, although the adversaries generated by above methods really work in the physical environments, they are still easily distinguishable to human vision in most cases because of their unnatural patterns.
%GAN can close the gap between the data manifolds of generated adversaries and real data. But due to the imperfections of GAN, the generated adversaries still look not natural enough.
\section{Harmonic Adversarial Attack Method}
In this section, we first introduce harmonic functions that are used as parametric models for constructing perturbations, and present an end-to-end procedure to learn the parameters. We also present methods to increase diversity of perturbasion by exploiting properties of harmonic functions.
\subsection{Why harmonic function?}
Harmonic functions can generate edge-free perturbations, from the perspectives of frequency domain analysis of images. 
%the definition of harmonic functions and give examples of harmonic perturbations, through which the readers can seize an intuitive understanding of why the perturbations generated by harmonic functions with a high magnitude can be edge-free.

First, in mathematics, let $f : U\to \mathbb{R}$ be a twice continuously differentiable function, where $U$ is an open subset of $\mathbb{R}^n$. If $f$ satisfies Laplace's equation everywhere on $U$, it is a harmonic function
~\cite{complex_gamelin}. The Laplace's equation is\par
\begin{equation}
\frac{\partial^2{f}}{\partial{x_1^2}}+\frac{\partial^2{f}}{\partial{x_2^2}}+\dots+\frac{\partial^2{f}}{\partial{x_n^2}}=0,
\end{equation}
which is also written as $\Delta f=0$. 

Considering the coordinate space of natural images. the region of harmonic functions should be in $\mathbb{R}^2$. According to Laplace's equation, it's easy to see that the sum of second-order derivatives with respect to variables equals to 0. This means the laplacian edge detector would never detect edges in the perturbations. We give some examples to show the differences between the laplacian maps of adversarial images generated using different methods in Fig.~\ref{fig:lap_cmp}.

In terms of the frequency domain, harmonic perturbations would not significantly affect the frequency components of natural images since they are very smooth in nature. This is quite different from noisy/edge-rich perturbations, which would add extra high frequency components in images' frequency domain ~\cite{signal_processing}.

\subsection{Generation of Harmonic Perturbation}
\label{subsec:harmonic_perturbation}
In this section, we will describe how to generate harmonic perturbations for natural images. The key is to construct a flexible harmonic function on the coordinate space of natural images. %We will give a mathematical formulation.\par 

Let $\mathcal{S}$ be a natural image set, and $\bm{I} \in \mathcal{S}$ denote a natural image with a size of $H\times W\times C$. We assume the space of pixels' coordinates in $\bm{I}$ to be a complex plane. Then the coordinate of one pixel $(x,y)$ can be represented as a complex number $z=x+yi$ where $x$ and $y$ are the real and imaginary part respectively. To define a universal framework for images with different size in $\mathcal{S}$, we normalize the coordinate space for all images into a zero-centered unit square, i.e. normalize both $x$ and $y$ into $[-1,1]$. %This is easily implemented in many scientific computing library\footnote{x = numpy.linspace(-1, 1, W), y=numpy.linspace(-1, 1, H).\\ W and H denote the width and height of one image.}, e.g. Numpy~\cite{numpy}.\par

To be compatible with mathematical theories, we assume the coordinate space (complex plane) of natural image is continuous instead of discrete. In the coordinate space, we define a complex function as $f(z;\theta)=f(x+yi;\theta)=u(x,y;\theta)+iv(x,y;\theta)$, where $\theta$ denotes the parameters in $f$. For example, a quadratic polynomial functions can be denoted as $f(z;\theta)=\theta_1 z^2+\theta_2 z+\theta_3$, where $\theta=\{\theta_1,\theta_2,\theta_3\}$. We take $f(z)$ to denote $f(z;\theta)$ for simplicity in the following parts. %As shown later, with one important restriction, we can derive a family of harmonic functions from the defined complex function $f(z)$.
  
Preliminarily, we give three properties which are crucial for deriving harmonic functions from $f(z)$.
%In fact, there are two important conditions we can use, with which we can derive a family of harmonic functions from $f(z)$.

\vspace{1mm}
\noindent \textbf{\emph{Property 1}:} \emph{If $f(z)$ satisfies the Cauchy-Riemann equations, it is an analytic function} ~\cite{complex_gamelin}.

\vspace{1mm}
\noindent \textbf{\emph{Property 2}:} \emph{ If $f(z)=u+iv$ is analytic in a region, then $u$ and $v$ satisfy Laplace's equation in the region, i.e. $u$ and $v$ are harmonic functions}~\cite{complex_gamelin}.

\vspace{1mm}
\noindent \textbf{\emph{Property 3}:} \emph{The linear combination of analytic functions is still an analytic function.}

\vspace{2mm}
According to \textbf{\emph{Property 1}}, we define $f(z)$ to satisfy the Cauchy-Riemann equations in order to make sure that it is an analytic function. Then based on \textbf{\emph{Property 2}}, we know the real and imaginary parts of $f(z)$ are harmonic functions and meanwhile they are conjugate~\cite{math_physics}.
In fact it has been proved that some known functions are analytic~\cite{complex_gamelin} e.g. polynomial, trigonometric, exponential functions. Not limited, we can build any complex function with just guaranteeing it satisfying the Cauchy-Riemann equations. Then, we can directly take the real part or imaginary part of complex functions as harmonic function to generate harmonic perturbations. We term the selected harmonic function as $h(x,y)$.

Next, with the aim of making the input image adversarial, we generate harmonic perturbations based on $h(x,y)$ which is defined upon the coordinate space. We First normalize the range of the $h(x,y)$ into[-1,1] with $norm(h)=(h-h_{min})*2/(h_{max}-h_{min})-1$. Then we use a coefficient $\epsilon$ to control the scale of harmonic perturbations when added on input images.
\begin{equation}
\label{eq:1}
\bm{I}_{dis}=clip_{[0,255]}(\bm{I}+\epsilon*norm(h))
\end{equation}

$\bm{I}_{dis}$ is the generated distorted image, and $clip_{[0,255]}$ denotes clipping the intensity of pixels in $\bm{I}_{dis}$ into [0,255]. To make $\bm{I}_{dis}$ adversarial to the target model, elaborate tunings of the parameters in $h(x,y)$ are desirable. We take the adversarial learning strategy to learn the parameters, which will be introduced in Sec.~\ref{subsec:learning}.

\begin{figure}[t]
\centering
\subfloat[(a) Combination of the real parts of $f(z)=(x+iy)^2$ and $f(z)=(x+iy)^3$.]{
\includegraphics[width=0.45\textwidth]{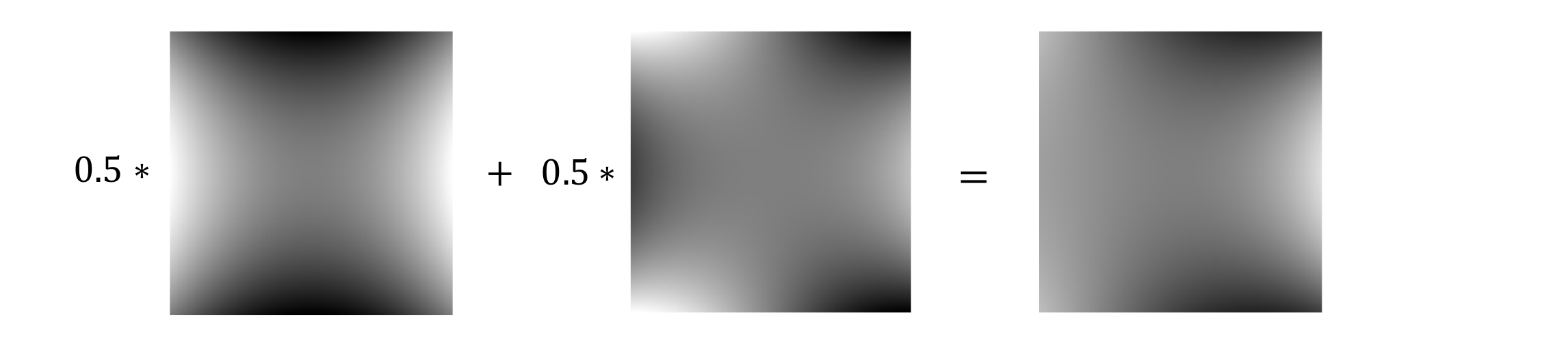}}\\
\subfloat[(b) Affine transformation example. Harmonic function is the real part of $f(z)=sin(x+iy)$. R denotes rotation. S denotes scaling. T denotes translation.]{
\includegraphics[width=0.45\textwidth]{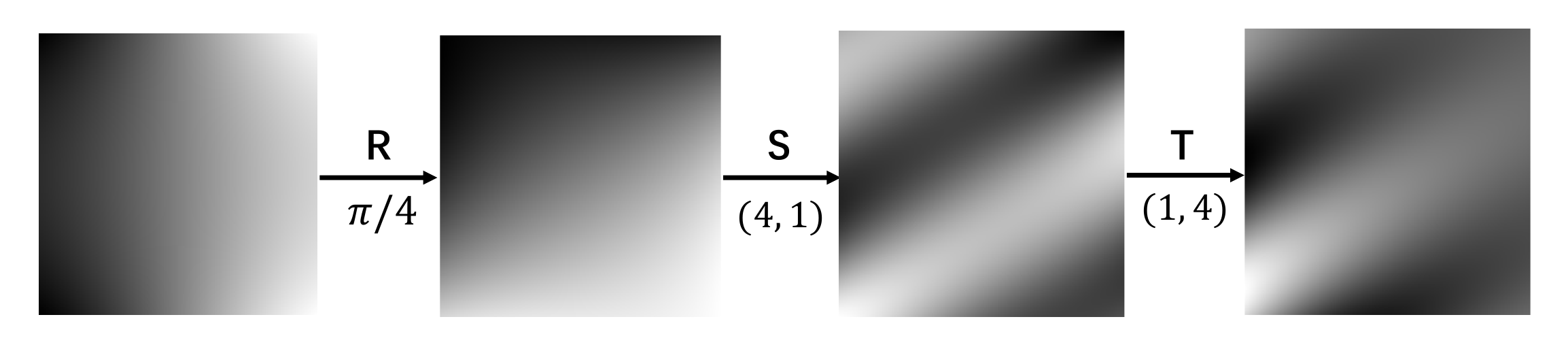}}
\caption{ Examples to show how expansion tricks work. }
\label{fig:aux_tricks}
\end{figure}

\subsection{Expansion Tricks for Perturbation Generation}
In addition, to generate more flexible and powerful harmonic perturbations, we suggest two expansion tricks, based on the properties of harmonic functions.

\subsubsection{Linear Combination of Harmonic Functions}
The selection of harmonic functions is the key to generate adversarial perturbations. We term the known harmonic functions as basic functions. With basic functions, we can construct more complicated functions according to \textbf{\emph{Property 3}}.

We select two known analytic functions: quadratic polynomial and sine function, which are marked as $f_{p}=u_p+iv_p$ and $f_{s}=u_s+iv_s$ respectively. It's flexible to select any basic functions, we here take these two functions as an example. The linearly combined function is $f_{c}=\alpha f_{p}+\beta f_{s}$, where $\alpha$ and $\beta$ are two learnable coefficients. It's easy to get the real part and imaginary part of $f_c$ are $u_c=\alpha u_p+\beta u_s$ and $v_c= \alpha v_p +\beta v_s$. And they both are harmonic functions according to \textbf{\emph{Property 2}} and \textbf{\emph{Property 3}}. We use $\mathcal{P}_{h}$ to denote the set of parameters in basic functions and learnable coefficients in the combined function. 

\subsubsection{Coordinate Space Affine Transformation}
We find applying affine transformations on the coordinate space can argument the harmonic function to generate powerful adversarial perturbations. We consider three transformations: rotation, scaling and translation. For any point $z=x+iy$ in the coordinate space (complex plane), three transformations are performed in turn.  
\begin{itemize}
\item \textbf{\emph{Rotation}}: Let $r$ denote the cosine value of the rotation angle, where $r \in [-1,1]$. 
\begin{equation}
\begin{bmatrix}x^{'}\\y^{'}\end{bmatrix}=\begin{bmatrix}r&-\sqrt{1-r^2}\\\sqrt{1-r^2}&r\end{bmatrix}*\begin{bmatrix}x\\y\end{bmatrix}
\end{equation}
\item \textbf{\emph{Scaling}}: Let $s_x$ and $s_y$ be the scale factors, where $s_x,s_y\in(0,10]$.
\begin{equation}
\begin{bmatrix}x^{''}\\y^{''}\end{bmatrix}=\begin{bmatrix}x^{'}\\y^{'}\end{bmatrix}*\begin{bmatrix}s_x\\s_y\end{bmatrix}
\end{equation}
\item \textbf{\emph{Translation}}: Let $t_x$ and $t_y$ be the translation distance, where $t_x, t_y\in[-1,1]$. 
\begin{equation}
\begin{bmatrix}x^{'''}\\y^{'''}\end{bmatrix}=\begin{bmatrix}x^{''}\\y^{''}\end{bmatrix}-\begin{bmatrix}t_x*s_x\\t_y*s_y\end{bmatrix}
\end{equation}
\end{itemize}

The parameters $\{r,s_x,s_y,t_x,t_y\}$ will be learned together with the parameters of the harmonic function. But differently, they are restricted in a narrow range. So we will give them a little learning rate during optimization. We mark the parameters set in affine transformations as $\mathcal{P}_{a}$.

To show how expansion tricks argument the generated perturbations, we give some examples in Fig~\ref{fig:aux_tricks}.

\subsection{Learning Strategy}
\label{subsec:learning}
\begin{figure}[t]
\centering
\includegraphics[width=0.45\textwidth]{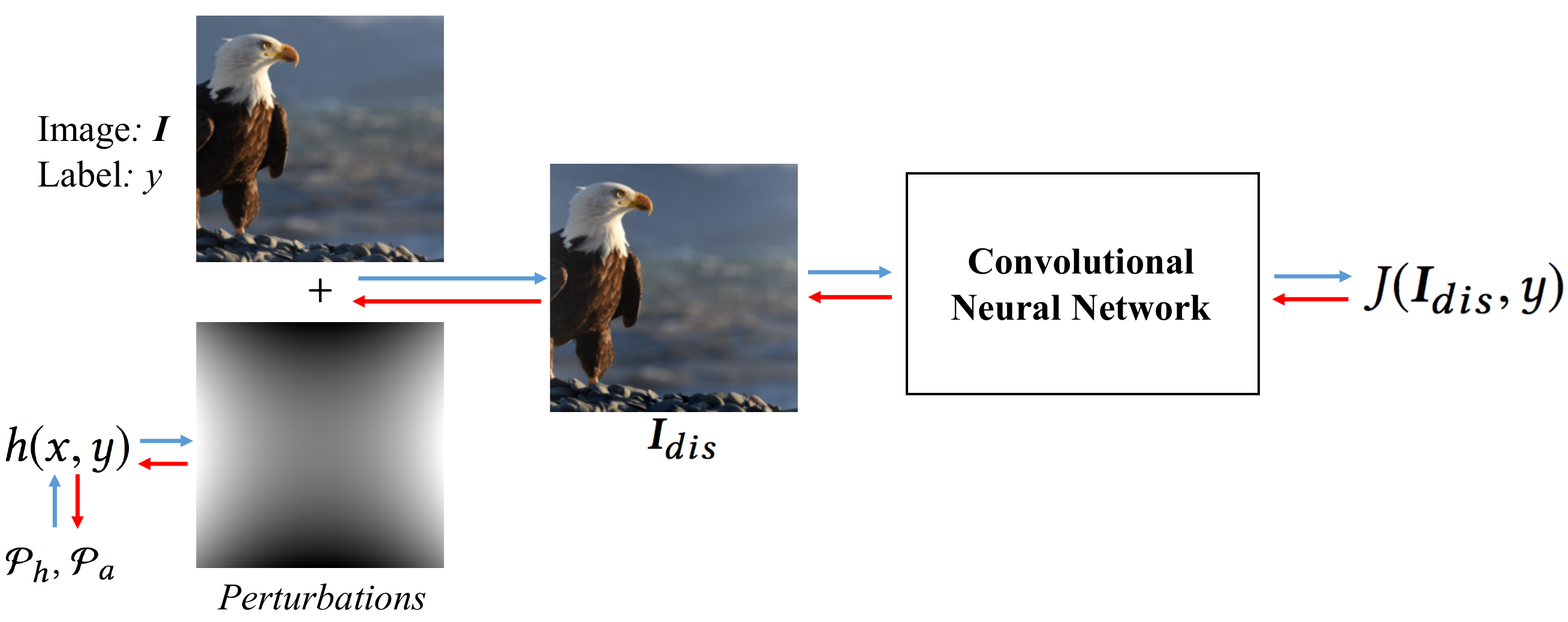}
\caption{ End-to-end workflow of HAAM. Blue arrows show the forward pass, and red arrows show the backward pass for updating parameters.}
\label{fig:HAAM_flow}
\end{figure}

The generation of harmonic perturbations depends on two groups of parameters: $\mathcal{P}_{h}$ and $\mathcal{P}_{a}$. We use the adversarial loss to learn these parameters. The learning process is illustrated in Fig.~\ref{fig:HAAM_flow}.

We assume the target model is a neural network classifier $F$ which predicts the logits for the given image. The number of classes is marked as $C$. $\bm{I}$ is one image in $\mathcal{S}$, its true class is $y$ where $y\in\{0,1,2,\dots,C-1\}$. For a classifier with a softmax prediction, when the class labels are integers the cross-entropy cost function equals the negative log-probability of the true class given the image. So we have the loss function as follows,
\begin{equation}
J(\bm{I}_{dis},y) = {\rm log}\ p(y|\bm{I}_{dis})
\end{equation}
where $\bm{I}_{dis}$ is generated from Eqn. (\ref{eq:1}). The above loss function is non-targeted adversarial attack loss, optimizing with which would decrease the predictive confidence of the given image with respect to its true class label.

We use the back-propagation algorithm to optimize the parameters. Since the parameters in affine transformations $\mathcal{P}_a$ are restricted in a narrow range, and parameters in the harmonic function $\mathcal{P}_h$ are freely adjustable, we set different learning rates ($lr$) for the optimization of parameters in these two modules. In our experiments, we set $lr=1e-1$ for the parameters in affine transformations and increase $10\sim 20$ times for parameters in harmonic functions. 

For the optimization procedure of the proposed algorithm please refer to \textbf{Algorithm} ~\ref{alg:1}. \par

\begin{algorithm}[h]
\caption{\small Optimization algorithm for HAAM } 
\raggedright
\hspace*{0.00in} {\bf Input:} natural image $\bm{I}$, label $y$, harmonic function $h$, target model $F$, number of iterations $T$, learning rate $lr_1$ for $\mathcal{P}_h$, $lr_2$ for $\mathcal{P}_a$ \\
\hspace*{0.0in} {\bf Output:} adversarial image $\bm{I}_{dis}$, indicator for adversary: $adv$\\
\hspace*{0.00in} {\bf Initialization:} $\mathcal{P}_{h}$, $\mathcal{P}_{a}$, $adv=False$ \\
\begin{algorithmic}[1]
\raggedright
\State $i=0$
\State $\bm{I}_{dis}=\bm{I}$
\While{$i<T$ and $F(\bm{I}_{dis})=y$}
	\State affine transform with $\mathcal{P}_a$ on coordinate space
    \State generate perturbations with $h(x,y)$ and $\mathcal{P}_h$
    \State generate $\bm{I}_{dis}$ according to Eqn. (\ref{eq:1})
    \State update $\mathcal{P}_h$: $\mathcal{P}_h=\mathcal{P}_h-lr_1\cdot\nabla_{\mathcal{P}_h} J(\bm{I}_{dis},y)$
    \State update $\mathcal{P}_a$: $\mathcal{P}_a=\mathcal{P}_a-lr_2\cdot\nabla_{\mathcal{P}_a} J(\bm{I}_{dis},y)$
    \State $i=i+1$
\EndWhile
\If{$F(\bm{I}_{dis})\ne y$}
	\State $adv=True$
\EndIf
\State \Return $\bm{I}_{dis}$, $adv$
\end{algorithmic}
\label{alg:1}
\end{algorithm}

\subsection{Gray-scale and Color Harmonic Perturbations}
\label{subsec:}
Since natural images have three color channels, we introduce two kinds of harmonic perturbations: gray-scale and color perturbations. A gray-scale perturbation means we learn a shared harmonic perturbations across all three channels. And a color perturbation means we learn separate perturbations for each channel of one image. %We give examples of adversarial images generated by these two ways in Fig.~\ref{fig:channel_cmp}. 
In experiments, we let HAAM-g denote HAAM with gray-scale perturbations and HAAM-c denote HAAM with color perturbations.

% \begin{figure}[h]
% \centering
% \subfloat [Predicted  label: chute]{
% \includegraphics[width=0.20\textwidth]{fig/haamg_dis.png}}
% \subfloat [Gray-scale perturbations]{
% \includegraphics[width=0.20\textwidth]{fig/haamg_e.png}}\\
% \subfloat [Predicted label: chute]{
% \includegraphics[width=0.20\textwidth]{fig/haamc_dis.png}}
% \subfloat [Color perturbations]{
% \includegraphics[width=0.20\textwidth]{fig/haamc_e.png}}
% \caption{ Examples to show the difference between gray-scale and color harmonic perturbations. Left column shows the adversarial images (true label: balloon), and there is almost no difference shown between the two adversarial images. Right column shows their corresponding harmonic perturbations. (Best viewed in color)}
% \label{fig:channel_cmp}
% \end{figure}
\section{Harmonic Perturbations vs. Natural Phenomena}
We find the the harmonic perturbations generated using special harmonic functions can simulate some natural phenomena or photographic effects in life. 

\begin{figure*}[t]
\centering
\subfloat[]{
\includegraphics[width=0.33\textwidth]{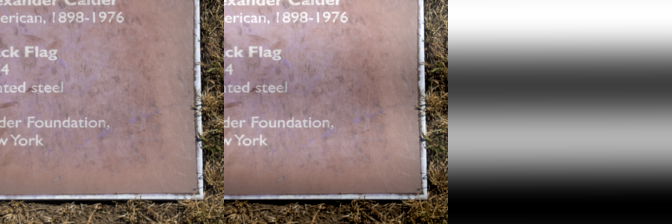}}
\subfloat[(a) Perturbations look like natural shadows.]{
\includegraphics[width=0.33\textwidth]{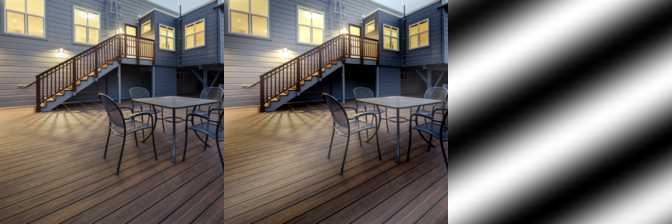}}
\subfloat[]{
\includegraphics[width=0.33\textwidth]{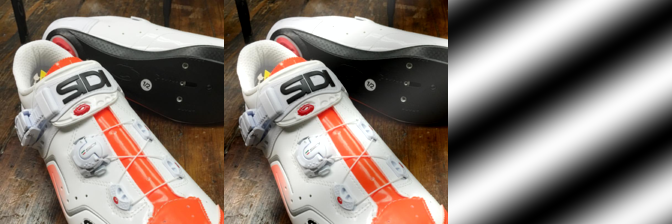}}\\
\subfloat{
\includegraphics[width=0.33\textwidth]{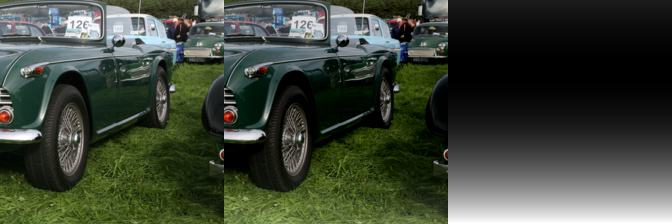}}
\subfloat[(b) Perturbations look like uneven exposure.]{
\includegraphics[width=0.33\textwidth]{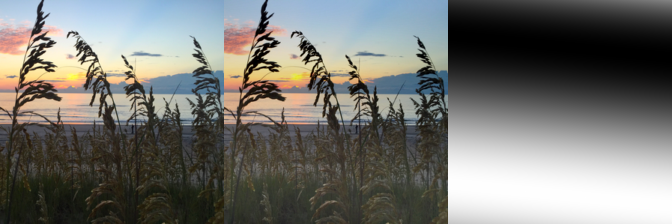}}
\subfloat{
\includegraphics[width=0.33\textwidth]{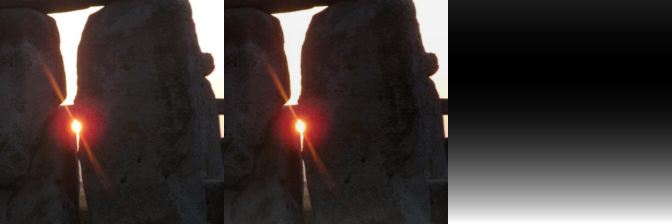}}\\
\subfloat[]{
\includegraphics[width=0.33\textwidth]{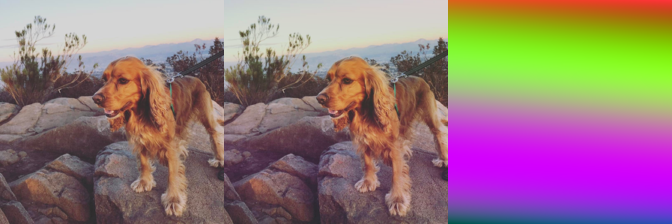}}
\subfloat[(c) Perturbations look like color adjusting.]{
\includegraphics[width=0.33\textwidth]{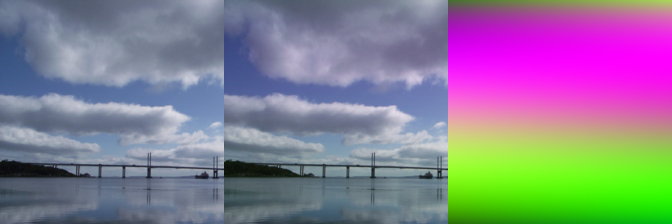}}
\subfloat[]{
\includegraphics[width=0.33\textwidth]{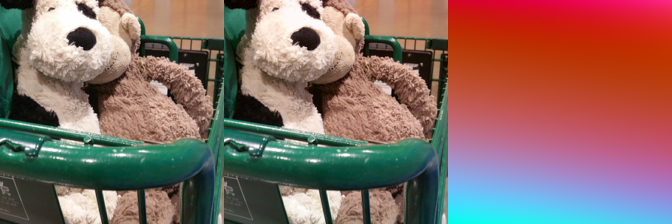}}
\caption{ Examples to show that harmonic perturbations look like natural phenomena or photographic effects. (a) shows gray-scale perturbations generated by harmonic functions from the real part of $sin(z)$ with affine transformations. (b) shows gray-scale perturbations generated by harmonic functions from the real part of polynomial functions with affine transformations. (c) shows color perturbations that look like colors adjusting using photo editors. (Best viewed in color)}
\label{fig:natural_pheno}
\end{figure*}

The adversarial images generated by harmonic functions of the real part of $f(z)=sin(z)$ show some kinds of stripe-like pattern. This pattern looks like the natural shadows or light in the scene when photos were shot. We give some examples in Fig~\ref{fig:natural_pheno} (a). In addition, if we use the the harmonic functions of the real part of polynomial analytic functions, the adversarial images look like showing some kinds of photographic effects, e.g. uneven exposure. Some adversarial images are shown in Fig~\ref{fig:natural_pheno} (b). If we generate channel-wise perturbations using combined harmonic functions, the adversarial images would look like with colors adjusting using photo editors. Examples are shown in Fig~\ref{fig:natural_pheno} (c).

In summary, the adversarial images generated using HAAM with specific harmonic functions can simulate some natural phenomena or photographic effects. In term of this point, the HAAM is quite different from previous methods which only generate noisy perturbations. HAAM can help to find some corner cases which possibly exist in our lives to cheat the target model, which in fact is useful in guiding design of data argumentation for training models, which will consequently make the model more robust in practice.      

\section{Experiments}
\label{sec:exp}

\begin{figure*}[t]
\centering
\subfloat{
\includegraphics[width=0.42\textwidth]{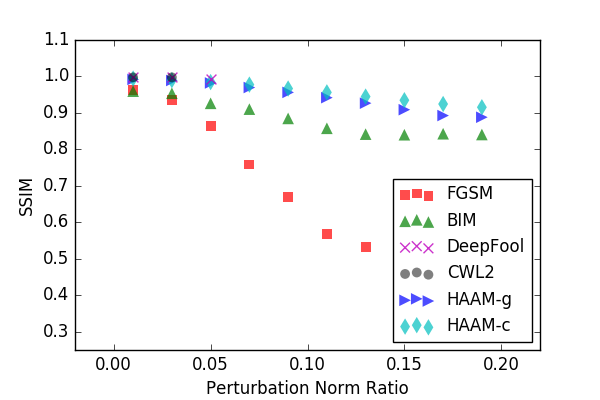}}
\subfloat{
\includegraphics[width=0.42\textwidth]{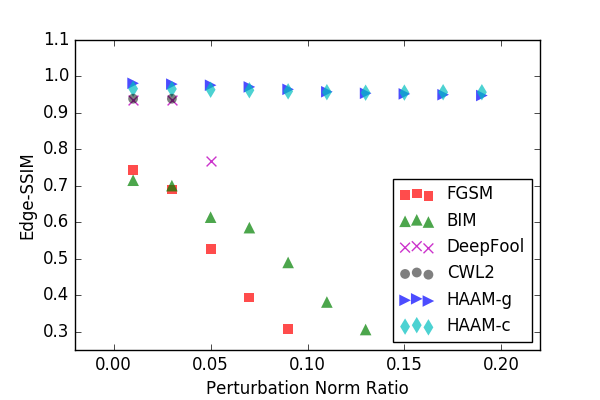}}
\caption{ SSIM and Edge-SSIM comparisons of different adversarial attack methods. (Zoom in to see details) }
\label{fig:pnr_ssim_essim}
\end{figure*}

We experiment on the Imagenet~\cite{imagenet} classification task. The dataset is from the competition `Defense Against Adversarial Attack' in NIPS 2017 ~\footnote{https://www.kaggle.com/c/nips-2017-defense-against-adversarial-attack}. There are dev-dataset (1000 images) and test-dataset (5000 images) released, we only use the test-dataset in our experiments. 

Multiple models trained on Imagenet dataset have been taken into account in our experiments, including Resnet50~\cite{resnet}, SqueezeNet~\cite{squeezenet}, VGG16~\cite{vgg} and Densenet121~\cite{densenet}. We take the Resnet50 as the main target model in the following experiments. All models are got from Pytorch official model zoo~\footnote{http://pytorch.org/docs/master/torchvision/models.html}.

\begin{table*}[!ht]  
\renewcommand\arraystretch{1.15}
\begin{minipage}{1\linewidth}
  \centering
  \footnotesize
  \caption{Transfer rate comparisons with different SSIM scores. The SSIM scores are in the range of [0.9, 1.0). We uniformly split the range into three buckets to calculate the transfer rate. The source model is Resnet50, and target models are SqueezeNet, VGG16 and Densenet121. There's no adversarial examples generated using DeepFool and CWL2 whose SSIM scores less than 0.967.}  
  \label{tab:ssim_tr}  
    \begin{tabular}{|c|c|ccc|c|}  
    \hlinew{1pt} 
    \makecell{SSIM}&Adv. Methods &SqueezeNet&VGG16&Densenet121& Average\cr \hlinew{1pt} 
    \multirow{5}{*}{[0.967, 1.0)}
    & FGSM       &0.349&0.208&0.195&0.251\cr\cline{2-6}
    & BIM       &0.350&0.239&0.247&0.279\cr\cline{2-6} 
    & DeepFool   &0.287&0.123&0.094&0.168\cr\cline{2-6}
    & CWL2       &0.459&0.188&0.088&0.245\cr\cline{2-6}
    & HAAM-g (ours)   &0.594&0.312&0.242&0.383\cr\cline{2-6}
    & HAAM-c (ours)   &\textbf{0.606}&\textbf{0.413}&\textbf{0.282}&\textbf{0.434}\cr\hlinew{1pt} 
    \multirow{5}{*}{[0.933, 0.967)}
    & FGSM       &0.424&0.318&0.341&0.361\cr\cline{2-6}
    & BIM       &0.425&0.409&\textbf{0.495}&0.443\cr\cline{2-6} 
    & DeepFool  	&-&-&-&-\cr\cline{2-6}
    & CWL2       &-&-&-&-\cr\cline{2-6}
    & HAAM-g (ours)       &0.625&0.371&0.284&0.427\cr\cline{2-6}
    & HAAM-c (ours)       &\textbf{0.652}&\textbf{0.432}&0.350&\textbf{0.478}\cr\hlinew{1pt} 
    \multirow{5}{*}{[0.900, 0.933)}
    & FGSM       &0.476&0.384&0.401&0.420\cr\cline{2-6}
    & BIM       &0.487&\textbf{0.503}&\textbf{0.608}&\textbf{0.533}\cr\cline{2-6} 
    & DeepFool  &-&-&-&-\cr\cline{2-6}
    & CWL2       &-&-&-&-\cr\cline{2-6}
    & HAAM-g (ours)       &0.673&0.381&0.381&0.478\cr\cline{2-6} 
    & HAAM-c (ours)       &\textbf{0.705}&0.314&0.339&0.453\cr\hlinew{1pt} 
    \end{tabular}  
\end{minipage}
\begin{minipage}{1\linewidth}
  \centering  
  \small
  \caption{Transfer rate comparisons with different Edge-SSIM scores. The Edge-SSIM scores are in the range of [0.8,1.0). We uniformly split the range into three buckets to calculate the transfer rate. The source model is Resnet50, and target models are SqueezeNet, VGG16 and Densenet121.}  
  \label{tab:essim_tr}  
    \begin{tabular}{|c|c|ccc|c|}  
    \hlinew{1pt}   
    \makecell{Edge-SSIM}&Adv. Methods &SqueezeNet&VGG16&Densenet121& Average\cr \hlinew{1pt} 
   \multirow{5}{*}{[0.933, 1.0)}
    & FGSM      &0.353	&0.184	&0.172	&0.236\cr\cline{2-6}
    & BIM       &0.347	&0.196	&0.188	&0.247\cr\cline{2-6} 
    & DeepFool   &0.292&0.122&0.092&0.169\cr\cline{2-6}
    & CWL2       &0.501&0.216&0.092&0.270\cr\cline{2-6}
    & HAAM-g (ours)   &0.620&\textbf{0.357}&0.286&0.421\cr\cline{2-6}
    & HAAM-c (ours)   &\textbf{0.637}&0.304&\textbf{0.408}	&\textbf{0.450}\cr\hlinew{1pt} 
    \multirow{5}{*}{[0.867, 0.933)}
    & FGSM       &0.329	&0.201	&0.344	&0.291\cr\cline{2-6}
    & BIM       &0.348	&0.231	&\textbf{0.391}	&0.323\cr\cline{2-6} 
    & DeepFool  	&0.283	&0.133	&0.165	&0.194\cr\cline{2-6}
    & CWL2       &0.402&0.150&0.086&0.213\cr\cline{2-6}
    & HAAM-g (ours)  &\textbf{0.690}&\textbf{0.386}	&0.317	&\textbf{0.464}\cr\cline{2-6}
    & HAAM-c (ours)       &0.620	&0.325&0.319&0.421\cr\hlinew{1pt} 
    \multirow{5}{*}{[0.80, 0.867)}
    & FGSM       &0.359	&0.245	&0.240	&0.281\cr\cline{2-6}
    & BIM       &0.363	&0.272	&0.311	&0.315\cr\cline{2-6} 
    & DeepFool  	&0.272	&0.100	&0.093	&0.155\cr\cline{2-6}
    & CWL2       &0.342&0.104&0.082&0.176\cr\cline{2-6}
    & HAAM-g (ours)       &\textbf{0.667}	&\textbf{0.394}	&\textbf{0.364}	&\textbf{0.475}\cr\cline{2-6} 
    & HAAM-c (ours)       &0.617	&0.209	&0.252	&0.359\cr\hlinew{1pt} 
    \end{tabular}  
\end{minipage}
\end{table*}

We make comparisons to existing adversarial attack methods from two aspects: visual quality and transferability.

Two metrics are selected to measure the visual quality.
\begin{itemize}
\item \textbf{SSIM}~\cite{ssim}: SSIM is one full-reference image quality assessment (IQA) method. It measures how much quality degradation of distorted image when compared to the reference image. Different from PSNR which is another one full-reference IQA method, the measurement of SSIM is more consistent with human vision. A higher SSIM score indicates a small quality degradation. 
\item \textbf{Edge-SSIM (ESSIM)}: ESSIM is our proposed metric, and it's also a full-reference quality assessment metric. ESSIM is to apply SSIM on the laplacian maps of the distorted image and the reference image. It measures how much distortions caused on the edge map. Since human vision is sensitive to edges in natural images, this metric is also very important for visual quality measurement. 
\end{itemize}

Transferability is measured by the transfer rate.
\begin{itemize}
\item \textbf{Transfer Rate (TR)}~\cite{adv_scale}: The transfer rate is calculated as follows,
\begin{equation}
\label{eq:tr}
TR=\frac{|S_t|}{|S_s|}
\end{equation}
where $|S_s|$ denotes the number of adversarial images generated by the source model, and $|S_t|$ denotes the number of adversarial images that successfully cheat the target model in $S_s$. $TR$ can reflect the transferability of the adversarial samples generated by one adversarial method. A higher TR means better transferability.
\end{itemize}

FGSM~\cite{FGSM}, BIM~\cite{BIM}, DeepFool~\cite{deepfool} and CWL2~\cite{cwl2} are considered for comparison in our experiments, and they all are classical and effective adversarial methods. All methods including our HAAM are implemented with Pytorch framework. FGSM, BIM and CWL2 are implemented with referring to their Tensorflow version in Cleverhans~\cite{cleverhans}, and the implementation of DeepFool is from the authors~\footnote{https://github.com/LTS4/DeepFool}.  It's worth noting that in FGSM, BIM and HAAM, there is a hyper-parameter $\epsilon$ to adjust the maximum magnitude ($L_\infty$) of generated perturbations. We select $\epsilon=\{1,2,4,8,16,24\}$ for FGSM, BIM and HAAM. For DeepFool and CWL2, there's no $\epsilon$ to control the magnitude of perturbations, so we just run it one time to generate adversarial examples on the testing dataset. The generated adversarial examples of all four adversarial methods will be used for all the remaining experimental analyses.

\subsection{Visual Quality Comparison}
\label{subsec:visual_cmp}
\subsubsection{Objective Comparison}
As highlighted in our paper, the adversarial examples generated using HAAM usually have a good visual quality on both the RGB color space and edge space. To make a fair comparison to other adversarial methods with respect to the visual quality, we will show the comparisons among several adversarial methods based on SSIM and Edge-SSIM metrics under the same perturbation magnitude. The magnitude of perturbation is measured by Perturbation Norm Ratio (PNR) metric ~\cite{deepfool} defined as $PNR=\|r(\bm{I})\|/\|\bm{I}\|$, where $r(\bm{I})$ denote the adversarial perturbations generated for image $\bm{I}$. This metric measures the magnitude of perturbations. A higher PNR indicates larger changes in the intensity of images.

We only consider the adversarial examples with their PNR values in the range of $(0,0.2]$, because PNR values of over 99\% of adversarial examples of each method are in this range. 
To compare the visual quality under the condition of same PNR score, we first split the range of (0,0.2] into 10 buckets uniformly (i.e. $bucket_1:(0,0.02], bucket_2:(0.02,0.04],\dots,bucket_{10}:(0.18,0.2]$). Then we put the adversarial examples into different buckets according to their PNR scores. We display the mean SSIM and center PNR value in each bucket in Fig.~\ref{fig:pnr_ssim_essim} (a). Similarly for ESSIM and PNR in Fig.~\ref{fig:pnr_ssim_essim} (b).

It's shown that the PNR values of adversarial examples generated using DeepFool and CWL2 are in a narrow range, and they show competitive visual qualities when compared to HAAM with respect to SSIM metric. But for Edge-SSIM metric, it's easy to see the DeepFool and CWL2 show worse laplacian map quality than HAAM. This indicates that although a slight noisy perturbation will not significantly affect the SSIM metric, but it inevitably brings changes in the edge space of natural images. Compared to FGSM and BIM, the adversarial examples generated using HAAM show significant advantages with respect to both the SSIM and Edge-SSIM metrics.

\subsubsection{Subjective Comparison}
Besides the comparison based on objective metrics, we further conduct subjective experiments to compare the visual quality of adversaries generated by HAAM-g and HAAM-c to other methods. We recruit 14 subjects including 10 males and 4 females in the subjective experiments, whose ages are in $19\sim25$.

Let's take the comparison between HAAM-g and FGSM as an example. First, we sample 100 pairs of adversarial images. In each pair, the two images are generated by HAAM-g and FGSM respectively from the same original image and they have very similar PNR values. Then we let subjects vote which image in the pair showing better visual quality. If one method gets the majority of 14 votes, it wins in this pair. Finally, we calculate the ratio of 100 image pairs won by HAAM-g, which is 0.96 in the following table. 

\begin{table}[!h]
\renewcommand\arraystretch{1.1}
  \centering
  \caption{Subjective visual quality comparisons between HAAMs and other adversarial methods.}
    \begin{tabular}{|c|c|c|c|c|}
    \hline
	& FGSM&BIM&DeepFool&CWL2\\
    \hline
    HAAM-g&0.96&0.99&0.54&0.50\\
    \hline
    HAAM-c&0.97&0.98&0.37&0.29\\
    \hline
    \end{tabular}
	\label{tab:sub_cmp}
\end{table}

Similarly, we perform the same procedure for other pairs of comparison methods. The ratios of HAAM are listed in Table~\ref{tab:sub_cmp}. It's shown that HAAM-g and HAAM-c outperform other methods in most cases, except comparing HAAM-c to DeepFool and CWL2.

\subsection{Transferability Comparison}
\label{subsec:tr_cmp}
Transfer testing is one kind of black-box attack ~\cite{transferability}. In reality, the target model is usually not accessible to attackers. So it's a feasible way to use the adversarial examples generated by a substitute model to attack the target model. In our experiment, we take the Resnet50 as the source model to generate adversarial examples, then take all adversarial examples to attack SqueezeNet, VGG16 and Densenet121. We take the metric of $TR$ to measure transferability.

We compare the transferability of adversarial examples generated using different adversarial methods under the condition of same visual quality. The visual quality are measured by the SSIM and Edge-SSIM metric. With the statistic analysis, we find there's a huge difference among the SSIM or Edge-SSIM score distributions of four adversarial methods. Over 99\% of adversarial examples generated using HAAM, DeepFool and CWL2 own SSIM scores in the range of $[0.9, 1.0)$ and Edge-SSIM scores in the range of $[0.8,1.0)$. But the SSIM scores and Edge-SSIM scores of adversarial examples generated using FGSM and BIM nearly uniformly distribute in the range of $[0.5,1)$ and $[0,1)$ respectively. To make an efficient comparison, we only consider the adversarial examples with SSIM scores in the range of $[0.9,1.0)$ and Edge-SSIM scores in the range of $[0.8,1.0)$.  

We split the range of SSIM and Edge-SSIM scores into three buckets uniformly. Then assign adversarial examples into corresponding buckets according to their SSIM or Edge-SSIM scores. For each bucket, we calculate the transfer rate using Eqn. (\ref{eq:tr}).

Transfer rates of different adversarial methods are shown in Table~\ref{tab:ssim_tr} and Table~\ref{tab:essim_tr}. In terms of the SSIM metric, it shows a trend that adversarial examples with a lower SSIM score have higher transferability (high TR). HAAM outperforms other methods on the phase of high SSIM scores ($>0.933$), but underperforms BIM on the phase of low SSIM scores. This is because the growth of TR of HAAM is slower than BIM while the SSIM score decreasing. In terms of the Edge-SSIM metric, HAAM outperforms other adversarial methods on all phases of Edge-SSIM scores and on all target models. That means with a similar edge (laplacian) map quality, the adversarial examples from HAAM always have a higher transferability.

\subsection{Failure Cases Analyses}
\begin{figure}[h]
\centering
\subfloat[(a) HAAM-g ($\epsilon=24$)]{
\includegraphics[width=0.16\textwidth]{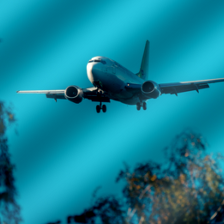}}
\subfloat[(b) HAAM-g ($\epsilon=24$)]{
\includegraphics[width=0.16\textwidth]{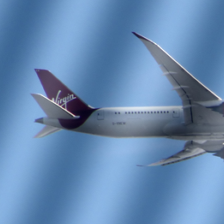}}
\subfloat[(c) HAAM-c ($\epsilon=16$)]{
\includegraphics[width=0.16\textwidth]{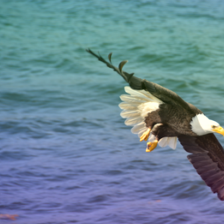}}
\caption{Adversarial examples that look unnatural to human vision. }
\label{fig:channel_cmp}
\end{figure}
We find that adversarial examples generated using HAAM  with a large $\epsilon$ show unnatural patterns to human vision at some special scenes e.g. sky, sea. These images are often with a simple flat background, which will highlight the unnaturalness of perturbations.

So, we suggest that attacking these kinds of images with HAAM should choose small $\epsilon$ and meanwhile selecting the harmonic functions without regular pattens (e.g. stripe-like pattern with sine). Moreover, gray-scale perturbations are more suitable for this scenario than color perturbations.
\section{Conclusion}
In this paper, we propose a new adversarial attack method - HAAM. Different from previous adversarial methods, HAAM generate edge-free perturbations, that are less disruptive to human vision compared to noisy/edge-rich perturbations. Experimentally, we find the adversarial examples generated by HAAM strike a balance between the visual quality and transferability between models. In addition, we find the adversarial examples generated by HAAM can simulate some nature phenomena or real-life photographic effects, which can be useful for the improvement of current DNNs, such as designing data augmentations in order to make the models more robust in practice.

\bibliographystyle{aaai}
\bibliography{reference}

\begin{thebibliography}{}

\bibitem[\protect\citeauthoryear{Athalye and Sutskever}{2017}]{print3d_adv}
Athalye, A., and Sutskever, I.
\newblock 2017.
\newblock Synthesizing robust adversarial examples.
\newblock {\em arXiv preprint arXiv:1707.07397}.

\bibitem[\protect\citeauthoryear{Boas}{2006}]{math_physics}
Boas, M.~L.
\newblock 2006.
\newblock {\em Mathematical methods in the physical sciences}.
\newblock Wiley.

\bibitem[\protect\citeauthoryear{Brown \bgroup et al\mbox.\egroup
  }{2017}]{adv_patch}
Brown, T.~B.; Man{\'e}, D.; Roy, A.; Abadi, M.; and Gilmer, J.
\newblock 2017.
\newblock Adversarial patch.
\newblock {\em arXiv preprint arXiv:1712.09665}.

\bibitem[\protect\citeauthoryear{Carlini and Wagner}{2017}]{cwl2}
Carlini, N., and Wagner, D.
\newblock 2017.
\newblock Towards evaluating the robustness of neural networks.
\newblock In {\em Security and Privacy (SP), 2017 IEEE Symposium on},  39--57.
\newblock IEEE.

\bibitem[\protect\citeauthoryear{Clark and Storkey}{2015}]{clark2015training}
Clark, C., and Storkey, A.
\newblock 2015.
\newblock Training deep convolutional neural networks to play go.
\newblock In {\em International Conference on Machine Learning},  1766--1774.

\bibitem[\protect\citeauthoryear{Deng \bgroup et al\mbox.\egroup
  }{2009}]{imagenet}
Deng, J.; Dong, W.; Socher, R.; Li, L.-J.; Li, K.; and Fei-Fei, L.
\newblock 2009.
\newblock Imagenet: A large-scale hierarchical image database.
\newblock In {\em IEEE Conference on Computer Vision and Pattern Recognition},
  248--255.
\newblock IEEE.

\bibitem[\protect\citeauthoryear{Evtimov \bgroup et al\mbox.\egroup
  }{2017}]{trafficsign_adv}
Evtimov, I.; Eykholt, K.; Fernandes, E.; Kohno, T.; Li, B.; Prakash, A.;
  Rahmati, A.; and Song, D.
\newblock 2017.
\newblock Robust physical-world attacks on deep learning models.
\newblock {\em arXiv preprint arXiv:1707.08945} 1.

\bibitem[\protect\citeauthoryear{Fawzi, Fawzi, and
  Frossard}{2018}]{fawzi2018analysis}
Fawzi, A.; Fawzi, O.; and Frossard, P.
\newblock 2018.
\newblock Analysis of classifiers' robustness to adversarial perturbations.
\newblock {\em Machine Learning} 107(3):481--508.

\bibitem[\protect\citeauthoryear{Gamelin}{2003}]{complex_gamelin}
Gamelin, T.
\newblock 2003.
\newblock {\em Complex analysis}.
\newblock Springer Science \& Business Media.

\bibitem[\protect\citeauthoryear{Goodfellow \bgroup et al\mbox.\egroup
  }{2014}]{gan}
Goodfellow, I.; Pouget-Abadie, J.; Mirza, M.; Xu, B.; Warde-Farley, D.; Ozair,
  S.; Courville, A.; and Bengio, Y.
\newblock 2014.
\newblock Generative adversarial nets.
\newblock In {\em Advances in Neural Information Processing Systems},
  2672--2680.

\bibitem[\protect\citeauthoryear{Goodfellow, Shlens, and Szegedy}{2014}]{FGSM}
Goodfellow, I.~J.; Shlens, J.; and Szegedy, C.
\newblock 2014.
\newblock Explaining and harnessing adversarial examples.
\newblock {\em arXiv preprint arXiv:1412.6572}.

\bibitem[\protect\citeauthoryear{He \bgroup et al\mbox.\egroup }{2016}]{resnet}
He, K.; Zhang, X.; Ren, S.; and Sun, J.
\newblock 2016.
\newblock Deep residual learning for image recognition.
\newblock In {\em IEEE conference on Computer Vision and Pattern Recognition},
  770--778.

\bibitem[\protect\citeauthoryear{Hinton \bgroup et al\mbox.\egroup
  }{2012}]{hinton2012deep}
Hinton, G.; Deng, L.; Yu, D.; Dahl, G.~E.; Mohamed, A.-r.; Jaitly, N.; Senior,
  A.; Vanhoucke, V.; Nguyen, P.; Sainath, T.~N.; et~al.
\newblock 2012.
\newblock Deep neural networks for acoustic modeling in speech recognition: The
  shared views of four research groups.
\newblock {\em IEEE Signal Processing Magazine} 29(6):82--97.

\bibitem[\protect\citeauthoryear{Huang \bgroup et al\mbox.\egroup
  }{2017}]{densenet}
Huang, G.; Liu, Z.; Weinberger, K.~Q.; and van~der Maaten, L.
\newblock 2017.
\newblock Densely connected convolutional networks.
\newblock In {\em IEEE conference on Computer Vision and Pattern Recognition},
  volume~1, ~3.

\bibitem[\protect\citeauthoryear{Iandola \bgroup et al\mbox.\egroup
  }{2016}]{squeezenet}
Iandola, F.~N.; Han, S.; Moskewicz, M.~W.; Ashraf, K.; Dally, W.~J.; and
  Keutzer, K.
\newblock 2016.
\newblock Squeezenet: Alexnet-level accuracy with 50x fewer parameters and< 0.5
  mb model size.
\newblock {\em arXiv preprint arXiv:1602.07360}.

\bibitem[\protect\citeauthoryear{Krizhevsky, Sutskever, and
  Hinton}{2012}]{krizhevsky2012}
Krizhevsky, A.; Sutskever, I.; and Hinton, G.~E.
\newblock 2012.
\newblock Imagenet classification with deep convolutional neural networks.
\newblock In {\em Advances in Neural Information Processing Systems},
  1097--1105.

\bibitem[\protect\citeauthoryear{Kurakin, Goodfellow, and Bengio}{2016a}]{BIM}
Kurakin, A.; Goodfellow, I.; and Bengio, S.
\newblock 2016a.
\newblock Adversarial examples in the physical world.
\newblock {\em arXiv preprint arXiv:1607.02533}.

\bibitem[\protect\citeauthoryear{Kurakin, Goodfellow, and
  Bengio}{2016b}]{adv_scale}
Kurakin, A.; Goodfellow, I.; and Bengio, S.
\newblock 2016b.
\newblock Adversarial machine learning at scale.
\newblock {\em arXiv preprint arXiv:1611.01236}.

\bibitem[\protect\citeauthoryear{Moosavi~Dezfooli, Fawzi, and
  Frossard}{2016}]{deepfool}
Moosavi~Dezfooli, S.~M.; Fawzi, A.; and Frossard, P.
\newblock 2016.
\newblock Deepfool: a simple and accurate method to fool deep neural networks.
\newblock In {\em IEEE Conference on Computer Vision and Pattern Recognition},
  number EPFL-CONF-218057.

\bibitem[\protect\citeauthoryear{Papernot \bgroup et al\mbox.\egroup
  }{2016}]{distillation}
Papernot, N.; McDaniel, P.; Wu, X.; Jha, S.; and Swami, A.
\newblock 2016.
\newblock Distillation as a defense to adversarial perturbations against deep
  neural networks.
\newblock In {\em Security and Privacy (SP), 2016 IEEE Symposium on},
  582--597.
\newblock IEEE.

\bibitem[\protect\citeauthoryear{Papernot \bgroup et al\mbox.\egroup
  }{2017}]{cleverhans}
Papernot, N.; Carlini, N.; Goodfellow, I.; Feinman, R.; Faghri, F.; Matyasko,
  A.; Hambardzumyan, K.; Juang, Y.~L.; Kurakin, A.; and Sheatsley, R.
\newblock 2017.
\newblock cleverhans v2.0.0: an adversarial machine learning library.

\bibitem[\protect\citeauthoryear{Papernot, McDaniel, and
  Goodfellow}{2016}]{transferability}
Papernot, N.; McDaniel, P.; and Goodfellow, I.
\newblock 2016.
\newblock Transferability in machine learning: from phenomena to black-box
  attacks using adversarial samples.
\newblock {\em arXiv preprint arXiv:1605.07277}.

\bibitem[\protect\citeauthoryear{Rabiner and Gold}{1975}]{signal_processing}
Rabiner, L.~R., and Gold, B.
\newblock 1975.
\newblock Theory and application of digital signal processing.
\newblock {\em Englewood Cliffs, NJ, Prentice-Hall, Inc., 1975. 777 p.}

\bibitem[\protect\citeauthoryear{Shapley and Tolhurst}{1973}]{shapley1973edge}
Shapley, R., and Tolhurst, D.
\newblock 1973.
\newblock Edge detectors in human vision.
\newblock {\em The Journal of physiology} 229(1):165--183.

\bibitem[\protect\citeauthoryear{Sharif \bgroup et al\mbox.\egroup
  }{2016}]{sunglass_adv}
Sharif, M.; Bhagavatula, S.; Bauer, L.; and Reiter, M.~K.
\newblock 2016.
\newblock Accessorize to a crime: Real and stealthy attacks on state-of-the-art
  face recognition.
\newblock In {\em ACM SIGSAC Conference on Computer and Communications
  Security},  1528--1540.
\newblock ACM.

\bibitem[\protect\citeauthoryear{Simonyan and Zisserman}{2014}]{vgg}
Simonyan, K., and Zisserman, A.
\newblock 2014.
\newblock Very deep convolutional networks for large-scale image recognition.
\newblock {\em arXiv preprint arXiv:1409.1556}.

\bibitem[\protect\citeauthoryear{Socher \bgroup et al\mbox.\egroup
  }{2011}]{socher2011parsing}
Socher, R.; Lin, C.~C.; Manning, C.; and Ng, A.~Y.
\newblock 2011.
\newblock Parsing natural scenes and natural language with recursive neural
  networks.
\newblock In {\em International Conference on Machine Learning},  129--136.

\bibitem[\protect\citeauthoryear{Stevens}{2015}]{v1_cortex}
Stevens, C.~F.
\newblock 2015.
\newblock Novel neural circuit mechanism for visual edge detection.
\newblock {\em Proceedings of the National Academy of Sciences}
  112(3):875--880.

\bibitem[\protect\citeauthoryear{Wang \bgroup et al\mbox.\egroup }{2004}]{ssim}
Wang, Z.; Bovik, A.~C.; Sheikh, H.~R.; and Simoncelli, E.~P.
\newblock 2004.
\newblock Image quality assessment: from error visibility to structural
  similarity.
\newblock {\em IEEE transactions on image processing} 13(4):600--612.

\bibitem[\protect\citeauthoryear{Zhao, Dua, and Singh}{2017}]{natural_adv}
Zhao, Z.; Dua, D.; and Singh, S.
\newblock 2017.
\newblock Generating natural adversarial examples.
\newblock {\em arXiv preprint arXiv:1710.11342}.

\end{thebibliography}

\end{document}